\documentclass[10pt,twocolumn,letterpaper]{article}

\usepackage{wacv}              

\usepackage{graphicx}
\usepackage{amsmath}
\usepackage{amssymb}
\usepackage{booktabs}

\usepackage[pagebackref,breaklinks,colorlinks]{hyperref}

\usepackage[capitalize]{cleveref}
\crefname{section}{Sec.}{Secs.}
\Crefname{section}{Section}{Sections}
\Crefname{table}{Table}{Tables}
\crefname{table}{Tab.}{Tabs.}

\def\wacvPaperID{2675} 
\def\confName{WACV}
\def\confYear{2024}

\begin{document}


\title{ Impact of Label Types on Training SWIN Models with Overhead Imagery}

\author{Ryan Ford$^1$, Kenneth Hutchison$^1$, Nicholas Felts$^1$,  Benjamin Cheng$^1$, Jesse Lew$^2$, Kyle Jackson$^3$ \\
$^1$Pennsylvania State University, $^2$ New York University,  $^3$ National Geospatial-Intelligence Agency\\
{\tt\small rzf5260@psu.edu}, {\tt\small Kyle.L.Jackson@nga.mil}
}
\maketitle

\begin{abstract}
Understanding the impact of data set design on model training and performance can help alleviate the costs associated with generating remote sensing and overhead labeled data. This work examined the impact of training shifted window transformers using bounding boxes and segmentation labels, where the latter are more expensive to produce. We examined classification tasks by comparing models trained with both target and backgrounds against models trained with only target pixels, extracted by segmentation labels. For object detection models, we compared performance using either label type when training. We found that the models trained on only target pixels do not show performance improvement for classification tasks, appearing to conflate background pixels in the evaluation set with target pixels. For object detection, we found that models trained with either label type showed equivalent performance across testing. We found that bounding boxes appeared to be sufficient for tasks that did not require more complex labels, such as object segmentation. Continuing work to determine consistency of this result across data types and model architectures could potentially result in substantial savings in generating remote sensing data sets for deep learning.
\end{abstract}

\section{Introduction}

For image analysis tasks, such as classification, object detection, and semantic segmentation, deep learning models have shown exceptional potential and performance. Because of this performance, application of deep learning models has become pervasive in multiple imaging fields, including remote sensing \cite{c1,c2,c3 }. Expanding the use of deep learning for tasks involving overhead imagery requires generating labeled data sets because most state-of-the-art approaches utilize supervised learning, where we supply the model with both the data and labels when training. While automated methods for generating data set labels are being developed \cite{c4}, we manually generate most data sets using human labor, which incurs substantial costs in time and effort. It is imperative to understand the impact of data set design on these image analysis tasks to optimally balance the cost of data set generation relative to model performance.

\begin{figure}[t]
  \centering
   \includegraphics[width=0.9\linewidth]{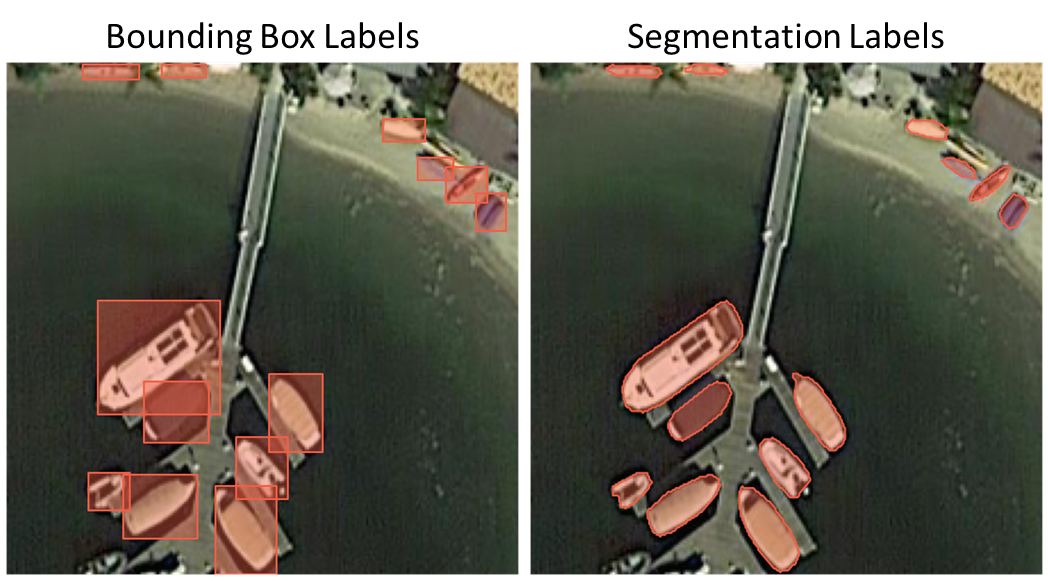}

   \caption{ Example of bounding box labels (left) and segmentation labels (right) for training object detection models. (Image Source: CVF DOTA Images)}
   \label{fig3}
\end{figure}

Label type is one aspect of data set design that must be considered during generation. Many overhead imagery data sets focus on object detection using simple bounding box labels \cite{c1,c5,c6,c7} because they are fast to generate and less costly. As target objects usually do not align with the image axes, bounding boxes rarely fit to target objects tightly, which results in background pixels in the labeled area, as shown in Figure \ref{fig3}. These background pixels may increase the potential for the model to correlate features in the background of the image to the target class, whether relevant or not. This learned background bias is common across deep learning domains focused on imagery and can negatively affect model generalization beyond the training set \cite{c8, c9, c10, c11}. In contrast, complex segmentation labels can be shaped to fit the target and avoid capturing background pixels at a higher production cost \cite{c12}. Additionally, the ability of segmentation labels to identify the target pixels in an image are useful in limiting background bias \cite{c13, c14}. Given this, it is worth determining if using complex labels in training deep learning models confers any performance benefit and if that benefit justifies the increased cost of production.

This paper examines the impact of label type on deep learning models to determine the performance difference between segmentation labels over bounding box labels, if any. We examined the utility of these label types in training both classification and objection detection models. These model types were chosen because they are used extensively in geospatial image analysis. Classification models are of interest because this examination allows analysis of their performance when the specific pixels pertaining to the target are known. Object detection models are a natural fit for this testing because they are inherently capable of being trained using either label type. For the classification model, we examined whether training with only the target pixels extracted via the segmentation labels resulted in better performance and less dependence on backgrounds than training with both the target and background pixels. For the object detection model, we examined training with either segmentation or bounding box labels and comparing the performance of each. Both tests utilized shifted window (SWIN) transformers \cite{c15}, selected for their notable performance in image analysis \cite{c16,c17}.

This work extends previous examinations of label type in overhead imagery \cite{c18}  by examining the utility of segmentation labels and the potential benefits they may impart. Further, this work extends the analysis to classification models, including examining potential approaches to reduce background bias. The outcomes of performing these analyses will assist in determining the utility of differing label types and aligning costs of data set development to the impact they have on model performance. The novel contributions of this work include:
\begin{itemize}
	\itemsep0em 
	\item An analysis of the impacts of using complex segmented versus simple bounding box labels when training machine learning models.
	\item Application of this analysis to tasks involving both object detection and classification in overhead, remotely-sensed imagery using vision transformer-based models.
	\item Examination of the utility of training classifiers using target pixels only, as extracted by segmentations, to reduce background bias of machine learning models.
\end{itemize}
\section{Related Work}
\label{sec:Related Work}

\subsection{Impact of Label Type on Model Training}

To the best of our knowledge, there is little published research on the impact of label type when training machine learning models. The use of label type is usually task dependent, with bounding boxes used in object detection and segmented labels in instance segmentation. While mean average precision (mAP) was the main evaluation metric used for both model types, the calculation of intersection over union (IOU) differed between tasks, thus their comparison was unequitable. Recently, researchers investigated the impact of label type when training object detection models by comparing the performance of convolutional neural network (CNN)-based models trained with various label types on the xView \cite{c6}, Dataset for Object detection in Aerial images (DOTA) 1.5 \cite{c19}, and FAIR1M \cite{c20} data sets \cite{c18}. The specific label types tested were centerpoints, image-aligned bounding boxes, and object-aligned bounding boxes. Comparison of performance across each of the tested models indicated that for tasks such as detection and counting training models with centerpoints showed comparable performance to full bounding boxes and oriented bounding boxes. They noted that some auxiliary tasks, such as determining the direction of an object’s travel, still required the more complex label types.

\subsection{Background Pixels and Spurious Correlations}

Deep learning models focused on image tasks have potential for developing spurious correlations between background and target pixels, resulting in poor generalization \cite{c8,c9,c10,c11,c21,c22}. For example, Lapuschkin, et al \cite{c10}. examined classifiers focused on both individual images and in aggregate using layer-wise relevance propagation and spectral relevance analysis. Researchers identified classifiers to find spurious correlations to background pixels to assist in classification, such as looking for the presence of water or sky in the imagery to classify boats or planes. Correlations can also arise from properties of the imagery that are imperceptible by human observers, such as noise \cite{c23, c24, c25} or from patterns that are indistinguishable by humans \cite{c26}, which are capable of resulting in high-confidence but incorrect predictions.

Various approaches were used to limit the development of spurious correlations in deep learning models. These approaches can be categorized as methods focused on augmenting training data, generating training data, or modifying the used model. Data augmentation approaches focused on modifying the target and background pixels within the data set. Augmentation approaches included randomly obscuring the data \cite{c27}, modifying the visual appearance of the target or background \cite{c9, c21, c28}, or using target and background pixels to generate contrastive examples \cite{c29}. Similar to data augmentation, data generation aimed to generate new data for training to increase the variability of the training data set, often with generative adversarial networks \cite{c30}. Model modification adapted the model architecture to limit the learning of spurious correlations. Examples of this approach include using ensembles that were trained to be robust to individual data perturbations \cite{c31} and supplementing the training process with additional information to indicate/prime the model to understand the location of the target pixels \cite{c13}. In determining the effects of label selection on model training, our work examined if the use of complex labels bound to the target can reduce the impact of spurious correlations from background pixels. In the case of the classification task, the segmented labels were used to extract the target pixels from the background, allowing training on the target pixels alone. In the case of the object detection task, using segmented labels shaped to the target acted as an indicator of target pixels when training the model.

\section{Methods}
\subsection{Models}
\subsubsection{Classification Model}

We used a simple SWIN block classifier written in TensorFlow \cite{c32} for the classification model in this experiment. The input layer ingested 128x128 px three-band (RGB) images in batches of 100. Our model used a patch size of (2,2), 4 attention heads, an embedding dimension of 64, a multilayer perceptron (MLP) size of 512, a window size of 4, and a window shift size of 2. We used an Adam optimizer with a learning rate of 1E-3 and trained our classification model with an 80 percent/20 percent train/validation split. We allowed the model to run for up to 150 epochs and stopped early if there was no improvement in validation loss for 10 epochs. For evaluation, we used only the weights with the lowest validation loss.

\subsubsection{Object Detection Model}

We used a Mask R-CNN model with a tiny SWIN (SWIN-T) backbone consisting of four SWIN blocks \cite{c33}, implemented in PyTorch \cite{c34} for the object detection model. The object detection model ingested 225x225 px three-band (RGB) imagery with labels in the COCO format \cite{c35}. We used the repository default hyper-parameters for testing. This included an embedding dimension of 96, a window size of 7, an MLP ratio of 4, and a varying number of attention heads with each block (3, 6, 12, and 24). These values are consistent with the parameters used in the original SWIN paper \cite{c15}. We trained the object detection models with an AdamW classifier and a learning rate of 5e-5. Models trained for a maximum of 15 epochs. We determined this value to be just beyond the point where the model improvement began to level out, based on experimentation.

\subsection{Dataset}

This work used DOTA \cite{c5}, which is an object detection data set with 15 classes across 2,806 images. The DOTA data set was collected from multiple sources to capture a range of sensors and spatial resolutions. The collection locations for the imagery were not stated, but the authors stated they tracked the selection process to avoid duplication of data. The train and validation portions of the DOTA data set were freely available. The test portion was not publicly available because it was used in an evaluation server provided by the authors. In this work, we split the training portion 80 percent/20 percent into train and validation sets. We used the DOTA validation set to evaluate our models and it was not used in training.

The DOTA data set used bounding box annotations that were not sufficient for examining simple versus complex labels. As such, this work also employed the Instance Segmentation in Aerial Images Dataset (iSAID) \cite{c36}, which supplied segmented labels for the imagery provided in the DOTA data set. The iSAID data set generated labels using human annotators and they were not based on the original DOTA labels. This resulted in 655,451 labels across the 15 classes. The 15 classes are Small Vehicle (250816 examples), Large Vehicle (35111 examples), Ship (33661 examples), Plane (6022 examples), Storage Tank (5250 examples), Harbor (3256 examples), Swimming Pool (2373 examples), Tennis Court (1159 examples), Helicopter (607 examples), Basketball Court (236 examples), Roundabout (212 examples), Bridge (197 examples), Soccer Ball Field (109 examples), Ground Track Field (83 examples), and Baseball Diamond (80 examples). The iSAID data set labels were freely available and provided in the COCO annotation format \cite{c35}. 

\subsection{Data Pre-Processing}
\subsubsection{Classification Pre-Processing}

Because the DOTA/iSAID data set was inherently an object detection data set, it required pre-processing to be applicable to classification tasks. Both the training and validation sets were converted to classification chips based on the iSAID labels. We removed objects that exceeded the maximum dimension allowed by our classification model, 128x128 px. Any data that had an improper number of channels was filtered, allowing only RGB imagery.

\begin{figure}[t]
  \centering
   \includegraphics[width=0.7\linewidth]{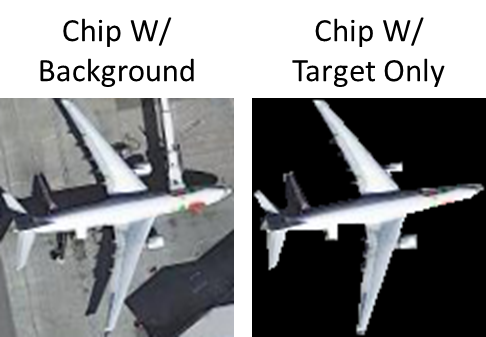}

   \caption{Example of training chip with background (left) and with background removed (right). (Image Source: CVF DOTA Images)}
   \label{fig2}
\end{figure}

As this experiment focused on examining the impact of simple labels versus complex labels, we generated two sets of training chips. The first set consisted of chips containing only target pixels, which were extracted using the iSAID segmentations. The chips were zero-padded to achieve the required image dimensions, avoiding the need for resizing while retaining aspect ratio. The second set consisted of chips containing both target and background pixels. In this case, we generated bounding boxes from the extent of the iSAID segmentations and extracted all pixels underneath. These chips were also zero-padded to the appropriate size to maintain aspect ratio. For regions of densely packed target objects, the chipping process made no effort to exclude nearby targets that were not the focus of the current chip. We also used these methods to generate chips for evaluation. Figure \ref{fig2} shows examples of each chip type.

We created two different compositions of the DOTA/iSAID data set to address the natural imbalance of classes in data sets consisting of overhead imagery. One composition used all classes and the other used only classes that had greater than 1000 examples. The latter composition specifically attempted to address the imbalance caused by classes with many examples. Further, this approach allowed us to determine if data set composition influences the effects of training with different label types and if we could replicate performance on different data sets. Only half of the DOTA/iSAID classes qualified for the majority class-only data set: storage tanks, large vehicles, small vehicles, planes, ships, swimming pools, harbors, and tennis courts. When training with the majority class-only data set, we used 10,000 examples per class. To achieve this, we randomly under sampled all classes with examples greater than 10,000 and over sampled all classes with less than 10,000 examples.

\subsubsection{Object Detection Pre-Processing}

Object detection pre-processing for the DOTA/iSAID data set was easier because the model supported COCO formatted labels by default. The DOTA imagery was larger than the input size required by the model and therefore chipped to size. In this chipping process, the imagery was resized to 1125x1125 px, padding the imagery beforehand as necessary, and divided into 5x5 sections. We removed any labels that intersected with the border of a chip because this process resulted in the labels having new coordinates in new images and updated the COCO label file with the changes.

We generated two different label files for the data set to explore the impact of different label types. The first label file consisted of the segmentation labels from the iSAID data set. The second label file consisted of bounding boxes generated based on the bounding shape of the segment labels. This represented the simple label data set with the bounding shapes covering both target and background pixels. Figure \ref{fig3} previously showed examples of both label types overlaying one of our training chips.

As with the classification model, we generated two different data set compositions. The first composition consisted of all classes in the data set and the second only used the two classes with the most examples. The rationale for the two-majority class data set was the same as that stated for the classification model: to evaluate label impact when training on a data set with smaller class imbalance. Given that the difference in the number of class examples between the classes in the two-majority data set was not as large as those in the majority only classification data set, the two-majority data set was neither over sampled nor under sampled. Additionally, as examples of both classes were found in a single object detection chip, this avoided the complication of needing to re-chip and resize the imagery.

\begin{figure*}[ht]
  \centering
   \includegraphics[width=0.6\linewidth]{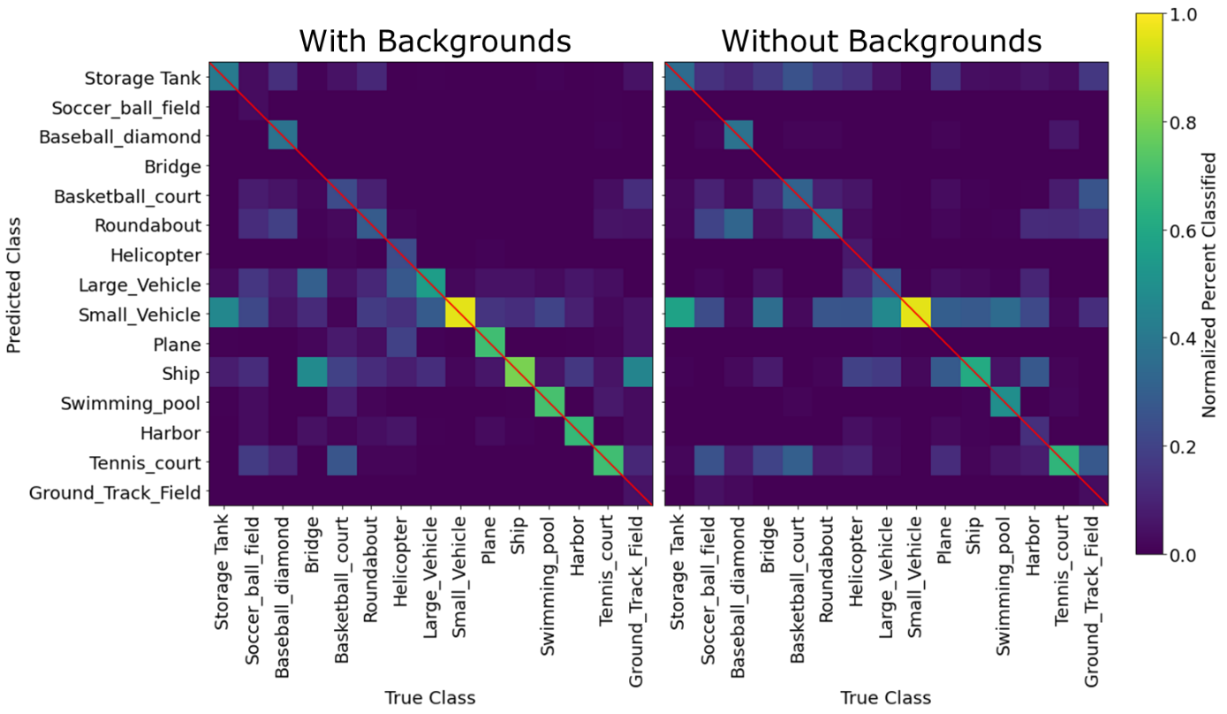}

   \caption{Confusion matrices for the model trained on the all-class data set with background and target pixels (left) and only target pixels (right). Cells under the red line indicate correctly classified targets.}
   \label{fig5}
\end{figure*} 

\subsection{Experiments}
\subsubsection{Classification Model}

The primary experiment performed for classification models was to examine the impact of training on images with backgrounds versus images without backgrounds. The classifier model trained on image chips with and without backgrounds from both the all-class and majority-class data sets. The data set pipeline normalized the images of both data sets by dividing by 255. Before normalizing the majority-class data set, the data set pipeline also performed a series of randomized augmentations on the data set using the keras preprocessing layers available in TensorFlow. These augmentations included random translation with height and width factor ranges of [0.0, 0.5] (i.e., up to half the image), both horizontal and vertical random flips, and random rotations with a factor of 1.0 (i.e., rotation up to 2 pi). Both the rotations and translations used the constant fill mode, which repeated the edge values for any points outside the image boundaries. Additional augmentations included random brightness adjustment with a factor range of [-0.1, 0.25] and random contrast adjustments with a factor range of [0.1, 0.5]. The purpose of these augmentations was to avoid overfitting to the over sampled classes. Because there was no over sampling in the all-class data set, we did not apply these augmentations. We repeated the model training three times, with the hyper-parameters, training set, and stopping criteria unchanged to attempt to quantify the variability in the training process. This resulted in three sets of model weights for evaluation. We used the accuracy, the individual class precision, and recall for evaluation. Reported results included aggregate accuracy of each of the three models. We reported the precision and recall of the best performing of the three models, based on their individual metrics, and the aggregate across all sets of weights and classes.

\subsubsection{Object Detection Model}

The primary experiment performed for the object detection model was an examination of the impact of training with simple labels versus complex labels. As stated in the preprocessing section, bounding box labels generated from the bounding shape of the segment labels represent our simple labels and the segment labels represent the complex labels. We trained two models, one with the simple labels and one with the complex labels, and compared them. We evaluated both the all-class and two-majority data sets, with no training differences between either.

We used mAP as the main evaluation metric for these tests. We determined these mAP values by IOU from bounding shapes as opposed to segments. This was a fairer comparison because this experiment was specifically testing object detection, not segmentation. The segment IOU may penalize results of the segmentation label-trained model if the prediction label only partially covered an object. Further, the larger size of the bounding box predictions from the simple model may have resulted in discarded predictions due to low IOUs though they correctly identified a target. We calculated mAP for the COCO standard IOU threshold (averaged across 0.5:0.05:0.95). In addition to mAP, we used individual class Average Precision (AP) to compare performance across classes.

\section{Results \& Discussion}
\subsection{Classification Model}

Classifier models trained on data with both background and target pixels outperformed models trained on data with only target pixels across all measured metrics. This result was consistent across both data set compositions used for testing. Table \ref{tab1} lists the performance metrics for both the models trained on both data set compositions. The mean class recall for the majority-class data set was consistently greater than that of the all-class data set. This indicates that the steps taken to balance the class distribution in the data set effectively reduced class biases. Using the majority class data set composition resulted in a lower accuracy than using all classes. This is likely due to the reduced performance on the classes with the greatest number of examples, because these classes also comprised the majority of the evaluation data set. Training with the all-class data set biased the model towards these classes. This bias was also likely the reason that the difference in accuracy between the background and no-background models was smaller for the all-class trained models than the majority-class models. Essentially, both models trained on the all-class data set were predisposed to favor the classes that appeared most often in the training set; therefore, the evaluation set also favored the classes that appeared most often.

\begin{table}
  \centering
  \begin{tabular}{c | c c c}
    \toprule
    Test & Acc. & Mean Class  & Mean Class  \\
    & & Recall & Precision \\
    \midrule
All (B) & $88.2 \pm 0.2$	&$44.9 \pm 2.3$	&$61.8 \pm 4.0$ \\
All (NB) & $81.2 \pm 0.8$	&$29.8 \pm 0.6$	&$30.9 \pm 2.3$ \\
Maj. (B) & $64.3 \pm 0.9$	&$64.3 \pm 0.8$	&$70.8 \pm 1.7$ \\
Maj. (NB) & $49.7 \pm 1.2$	&$49.7 \pm 1.2$	&$53.2 \pm 1.3$\\
    \bottomrule
  \end{tabular}
  \caption{Performance metrics for the all-class (All) and majority class (Maj.) classification models trained with backgrounds (B) and without backgrounds (NB). Values given are the aggregate performance of multiple sets of trained weights.}
  \label{tab1}
\end{table}

\begin{figure*}[t]
  \centering
   \includegraphics[width=0.8\linewidth]{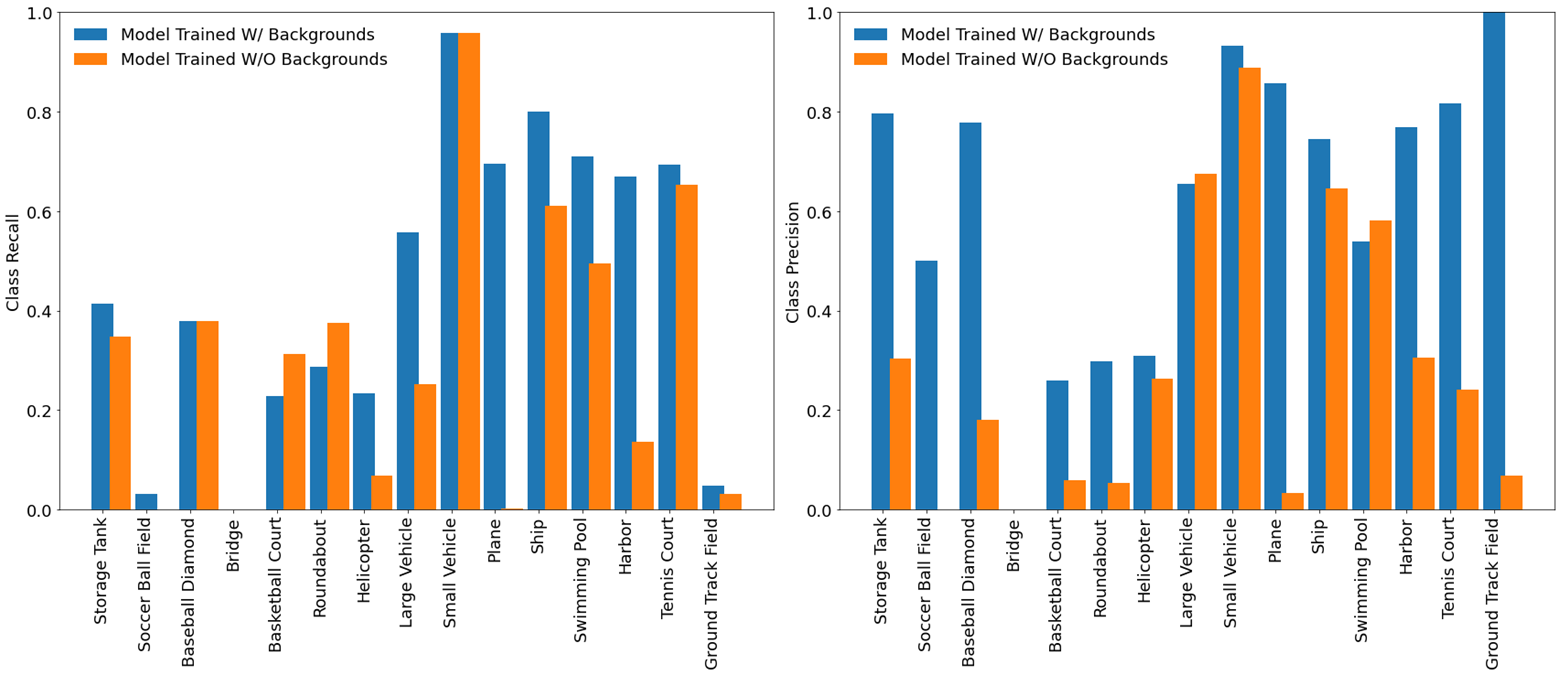}

   \caption{Class recall (left) and precision (right) for the models trained on the all-class data set.}
   \label{fig6}
\end{figure*}

Figure \ref{fig5} shows confusion matrices for both models trained on the all-class data set and Figure \ref{fig6} shows the individual class recall and precision. Both the confusion matrix and recall plots indicated the classifier trained with backgrounds performed best on the classes with the most examples, again an indicator of the model’s bias. The “With Background” confusion matrix also shows that this model often misclassified smaller classes as small vehicles, when small vehicles were the largest classes in the data set. Classes with few examples showed markedly poorer performance, either rarely or never classifying correctly.

The “Without Background” confusion matrix in Figure \ref{fig5} shows similar trends to those of the ”With Background” results. The best performing classes were also those with a majority of examples. Compared to the model trained with backgrounds, performance on these classes was notably degraded. Most classes were misclassified as small cars; however, misclassification as storage tank noticeably increased. This was indicated both in the “Without Background” confusion matrix and by comparing the precision for storage tanks in both models. Given that the storage tank class consisted of large objects that took up a large portion of the image chip, this may suggest that the model trained without backgrounds considered the background pixels present in the evaluation set as part of the target. This raises the question as to why many other large targets in the DOTA/iSAID data set, such as the various sports fields, did not become sources of common misclassification. This could potentially be due the storage tank class having more examples in the data set or the size of the storage tank objects being a better compromise between the small and large object classes.

\begin{figure*}[t]
  \centering
   \includegraphics[width=0.6\linewidth]{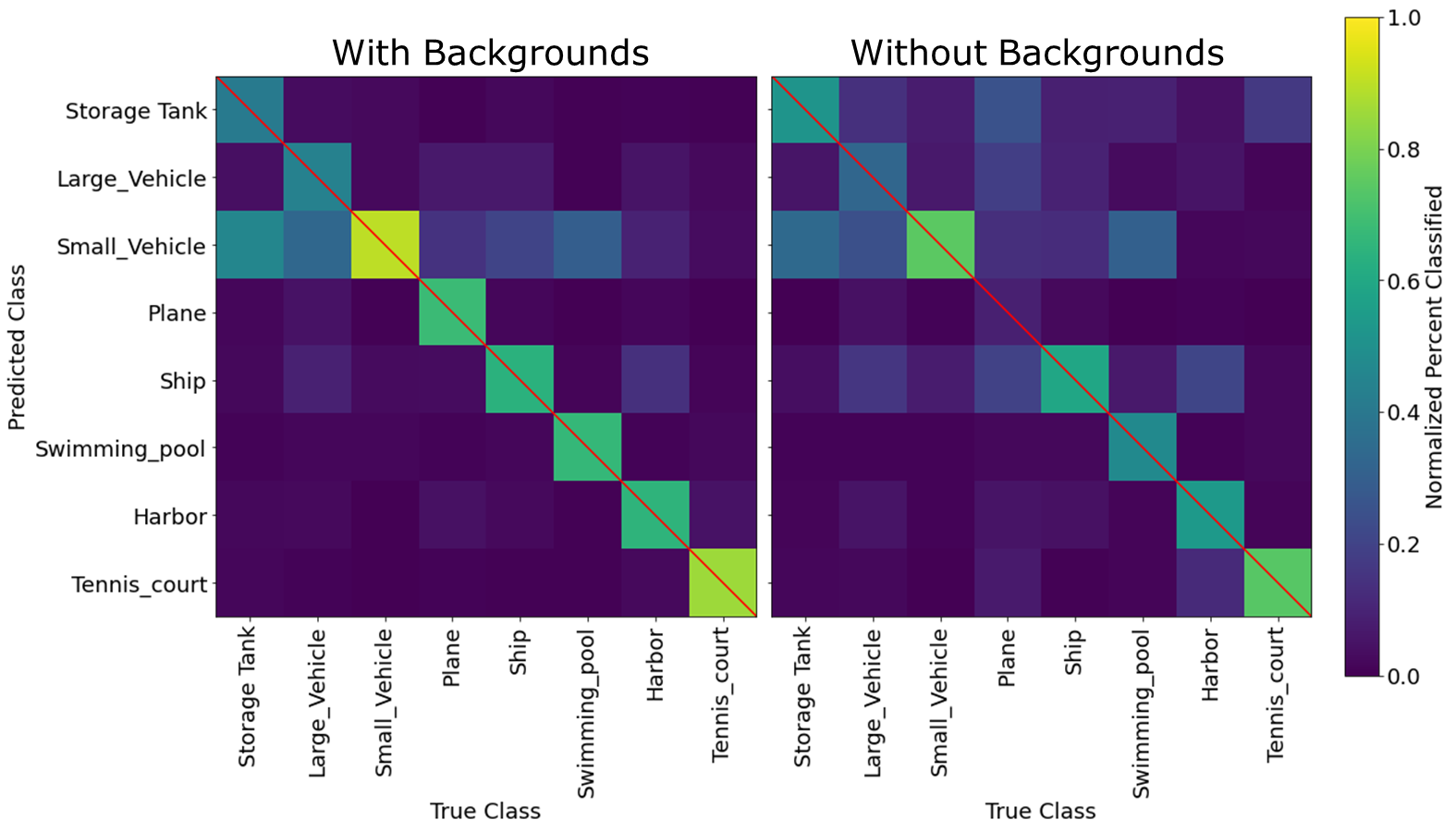}

   \caption{Confusion matrices for the model trained on the majority-class only data set with background and target pixels (left) and only target pixels (right). Cells under the red line indicate correctly classified targets.}
   \label{fig7}
\end{figure*}

\begin{figure*}[t]
  \centering
   \includegraphics[width=0.8\linewidth]{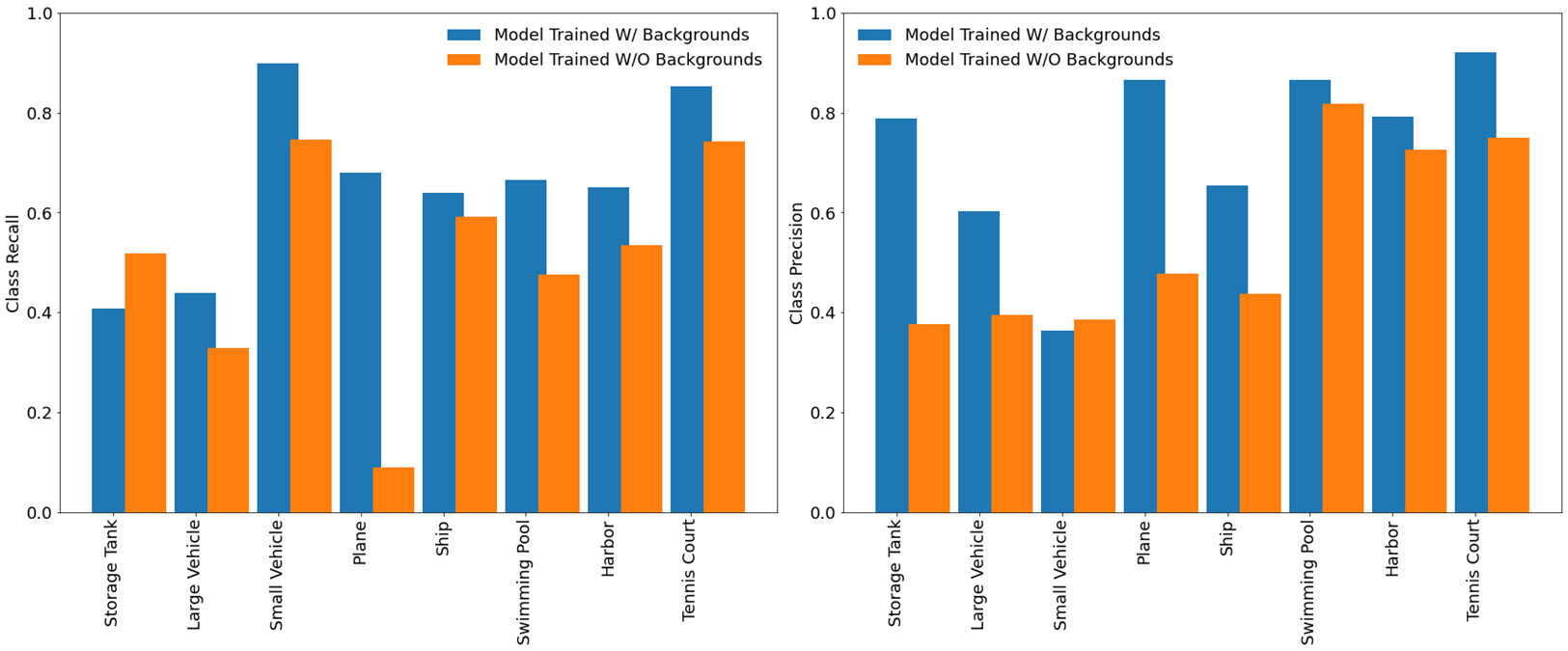}

   \caption{Class recall (left) and precision (right) for the models trained on the majority-class only data set.}
   \label{fig8}
\end{figure*}

Figures \ref{fig7} and \ref{fig8} present the confusion matrices and class-specific precision/recall for the majority-class only data set, respectively. The trends noted in the all-class data set models also were apparent in the results for the majority-class only models. The classes with the majority examples performed best, with the model most often misclassifying objects as the largest class in the data set. Because this data set composition was specifically limited to the largest classes, the improved class recall performance was expected. Again, as with the all-class data set, performance was lower for the model trained without backgrounds. Small vehicles continued to be the best performing class; however, storage tanks again became a more consistent cause of misclassification. The rationale for this increase was considered the same as for the all-class data set - that the model trained without backgrounds assumed the background pixels in the evaluation set were related to the target, making the target appear to be a large storage tank-like object.

Table \ref{tab2} gives further evidence that the decreased performance of the model trained with only target pixels resulted from conflating background and target pixels. Table \ref{tab2} shows the performance of both target-only trained models on the same evaluation set used for testing, but with the background pixels removed. The model’s performance was similar to, and sometimes surpassed, that of the models trained with backgrounds. This finding was in line with that of Kc, et al \cite{c21}., where models trained and evaluated on data with backgrounds removed outperformed models trained on the unmodified data. This indicated that training models with only the target pixels did not improve model generalization and resulted in models that were not trained with sufficient variability to represent real-world scenarios.

\begin{table}[h]
  \centering
  \begin{tabular}{c | c c c c}
    \toprule
    Test & Acc. & Mean Class  & Mean Class  \\
    & & Recall & Precision \\
    \midrule
All (NB) & $88.8 \pm 0.3$	&$47.1 \pm 1.2$	&$61.7 \pm 2.1$ \\
Maj. (NB) & $70.4 \pm 3.2$	&$70.4 \pm 3.2$	&$74.4 \pm 2.4$\\
    \bottomrule
  \end{tabular}
  \caption{Performance metrics for the all-class (All) and majority-class (Maj.) classification models without backgrounds (NB) on an evaluation set with only target pixels. Values given are the aggregate performance of multiple sets of trained weights.}
  \label{tab2}
\end{table}

\subsection{Object Detection Model}

\begin{table}[h]
  \centering
  \begin{tabular}{c | c c c c}
    \toprule
    Model & Seg. Model  & BBox Model  \\
    & mAP & mAP \\
    \midrule
All Classes & 0.239 & 0.236 \\
Two-Maj. Classes & 0.192 & 0.192 \\

    \bottomrule
  \end{tabular}
  \caption{Mean Average Precision for object detection models trained with simple and complex labels.}
  \label{tab3}
\end{table}

\begin{figure}[t]
  \centering
   \includegraphics[width=1.0\linewidth]{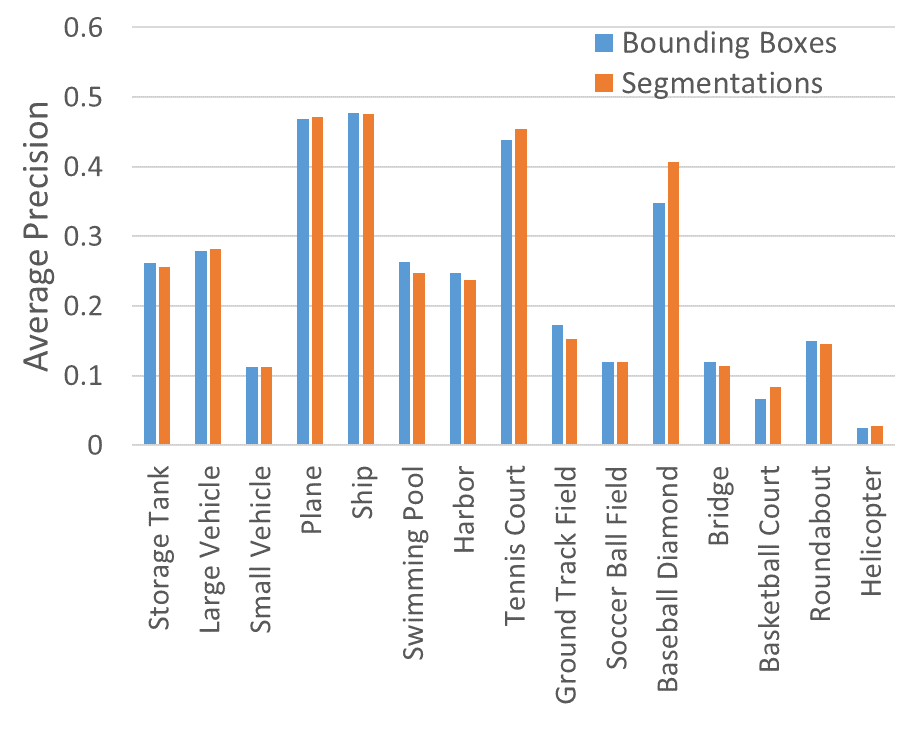}

   \caption{Individual class average precision for the object detection model trained with all classes. Blue shows results from the model trained with simple bounding boxes and orange shows results from the model trained with complex segmentation.}
   \label{fig10}
\end{figure}

Table \ref{tab3} shows the results of evaluating object detection models trained with simple bounding box labels and complex segmentation labels. For models trained with all classes, the mAP was nearly identical with a difference of only 0.3 percent in favor of the segmentation labels. Figure \ref{fig10} shows the individual class APs. For nearly all classes, the performance between the models was almost identical with differences of less than 1 percent. Performance on baseball diamonds showed the largest difference, nearly 6.0 percent lower for the bounding-box trained model. Baseball diamonds were the smallest class in the DOTA/iSAID data set, with only 80 and 36 examples in our training and evaluation set, respectively. Given the small set of examples in the evaluation set, this large difference in mAP was likely just the result of a few missed detections by the bounding box trained model. Overall, when training with the all-class data set, label type appeared to have little to no effect.

The results of performing the label test on the two-majority data set reinforced those of the all-class data set. Examining the mAP in Table \ref{tab3} showed identical performance when training with either label type. For the two-class model, the APs for small vehicles were 0.111 versus 0.112 and for large vehicles 0.273 versus 0.272, given as models trained on bounding boxes versus segmentations. These APs were nearly identical for models trained on both data set compositions even though the two-majority data set substantially reduced the number of classes. We expected to see performance increase due to reduced task complexity from only focusing on the classes with the greatest number of examples. This suggested other factors of the data and training may be limiting performance. One potential cause may be the relatively small size of objects in these classes and their appearance in the varying spatial resolutions of the imagery. The resampling process also could potentially exacerbate this effect. Regardless, the equal performance in this label test, and the previous test, is a strong indicator that the use of bounding boxes does not have a notable impact on object detection performance.

\subsection{Limitations}

While the results presented here indicate that there is no noticeable gain in performance when using complex segmentation labels over simple bounding box labels, this examination only explored a single configuration for testing. Further examination of label types should determine if these results are consistent. This work focused only on SWIN transformers. Testing should be replicated with other model types, including other vision transformers and convolutional neural networks. Examination of SWIN transformers could also be extended to other label types, such as centerpoints tested by Inder, et al \cite{c18}. Alternate approaches to training models and hyper-parameter tuning should also be considered. Researchers should also examine additional approaches to utilize segmentation labels, such as generating synthetic data with the known target pixels or using the segmentation labels to train the classifier with attention for the classification task. This analysis should extend beyond the DOTA/iSAID data sets to examine if these results are consistent across other spatial resolutions, spectral coverages, modalities, and other properties over which geospatial image data varies.

\section{Conclusion}

This work investigated whether using segmented labels is worth the increased cost of production over simpler label types. We compared the utility of segmentation labels and bounding box labels when training both SWIN transformer-based classifiers and object detection models. For classification models, we examined the utility of using data with only target pixels generated from the segmentation labels for training with only target pixels versus training with both target and background pixels. For object detection models, we compared models trained with either label type.

Examining classifiers indicated that models trained with both target and backgrounds consistently outperformed models trained with targets only. Examination of the performance of the classifier trained without backgrounds suggested that it considered the background pixels in the evaluation set related to the target. As such, we concluded that training with target pixels only data does not improve model performance and generalization. Results from training object detection models with varying label types indicated that the use of segmentation labels did not result in any performance improvement. Models showed only a slight difference in aggregate and individual class performance, but these differences could be a result of training variance. The overall selection of label type appeared to have no impact on the model.

For the models and training approaches tested, segmented labels did not provide an added benefit to model performance indicating that bounding boxes appear sufficient for tasks that did not require segmentations. These results, in conjunction with future additional tests, will help expand our knowledge of how deep learning models handle label types and background pixels, ideally resulting in greater generalizability and applicability of models and more effective development of training data sets.

\subsection*{Acknowledgement}
This work was supported by NGA contract HM0476-21D-0001. Any opinions, findings and conclusions or recommendations expressed in this material are those of the author(s) and do not necessarily reflect the views of NGA, DoD, or the US government. Approved for public release, NGA-U-2023-01872.

{\small
\bibliographystyle{ieee_fullname}
\bibliography{ford2023}
}

\end{document}